\documentclass[letterpaper, 10 pt, journal, twoside]{ieeetran}
\IEEEoverridecommandlockouts                              % This command is only needed if 
\usepackage{amsmath}
\usepackage{mathtools}
\usepackage[linesnumbered,ruled,vlined]{algorithm2e}
\usepackage{amsfonts}
\usepackage{xr-hyper}
\usepackage{hyperref}
\usepackage{amssymb}
\usepackage{graphics} % for pdf, bitmapped graphics files
\usepackage{epsfig} % for postscript graphics files
\usepackage{mathptmx} % assumes new font selection scheme installed
\usepackage{times} % assumes new font selection scheme installed
\usepackage[linesnumbered,ruled,vlined]{algorithm2e}
\usepackage{subcaption}
\usepackage[utf8]{inputenc}
\usepackage{graphicx}
\usepackage{bm}
\usepackage{xcolor}
\usepackage{enumitem}
\usepackage{cleveref}
\usepackage{acronym}
\usepackage{tabularx}
\usepackage{balance}
\setlist{nolistsep}
\newcommand{\textquote}[1]{\textit{#1}}
\newcommand{\action}[1]{\texttt{#1}}
\newcommand{\alfred}{ALFRED}
\newcommand{\dialfred}{DialFRED}

\title{\dialfred: Dialogue-Enabled Agents for Embodied Instruction Following}

\author{Xiaofeng Gao$^{1}$, Qiaozi Gao$^{2}$, Ran Gong$^{1}$, Kaixiang Lin$^{2}$, Govind Thattai$^{2}$, Gaurav S. Sukhatme$^{3}$%
\thanks{Manuscript received: February 24, 2022; Revised June 6, 2022; Accepted July 8, 2022.}%Use only for final RAL version
\thanks{This paper was recommended for publication by Editor Gentiane Venture upon evaluation of the Associate Editor and Reviewers' comments. This work was supported by Amazon Alexa AI.} %Use only for final RAL version
\thanks{$^{1}$Xiaofeng Gao and Ran Gong are with the Center for Vision, Cognition, Learning, and Autonomy, UCLA.
        {\tt\footnotesize \{xfgao, nikepupu\}@ucla.edu}}%
\thanks{$^{2}$Qiaozi Gao, Kaixiang Lin and Govind Thattai are with Amazon Alexa AI.
        {\tt\footnotesize \{qzgao, kaixianl, thattg\}@amazon.com}}%
\thanks{$^{3}$Gaurav S. Sukhatme is with Amazon Alexa AI and the Department of Computer Science, USC Viterbi School of Engineering. {\tt\footnotesize sukhatme@amazon.com,  gaurav@usc.edu}}

\thanks{Digital Object Identifier (DOI): see top of this page.}
}

\makeatletter
\newcommand*{\addFileDependency}[1]{% argument=file name and extension
  \typeout{(#1)}
  \@addtofilelist{#1}
  \IfFileExists{#1}{}{\typeout{No file #1.}}
}
\makeatother

\begin{document}

% Paper headers
\markboth{IEEE Robotics and Automation Letters. Preprint Version. Accepted JULY, 2022}
{Gao \MakeLowercase{\textit{et al.}}: \dialfred: Dialogue-Enabled Agents for Embodied Instruction Following} 

\maketitle
% \thispagestyle{empty}
% \pagestyle{empty}

%%%%%%%%%%%%%%%%%%%%%%%%%%%%%%%%%%%%%%%%%%%%%%%%%%%%%%%%%%%%%%%%%%%%%%%%%%%%%%%%
\begin{abstract}
Language-guided Embodied AI benchmarks requiring an agent to navigate an environment and manipulate objects typically allow one-way communication: the human user gives a natural language command to the agent, and the agent can only follow the command passively. We present \textbf{\dialfred}, a dialogue-enabled embodied instruction following benchmark based on the \alfred\ benchmark. \dialfred\ allows an agent to actively ask questions to the human user; the additional information in the user’s response is used by the agent to better complete its task. We release a human-annotated dataset with 53K task-relevant questions and answers and an oracle to answer questions. To tackle \dialfred, we propose a questioner-performer framework wherein the questioner is pre-trained with the human-annotated data and fine-tuned with reinforcement learning.
Experimental results show that asking the right questions leads to significantly improved task performance. We make \dialfred\ publicly available and encourage researchers to propose and evaluate their solutions to building dialog-enabled embodied agents: \href{https://github.com/xfgao/DialFRED}{https://github.com/xfgao/DialFRED}
\end{abstract}


\section{Introduction}
\IEEEPARstart{R}{obot} assistants need to understand natural language and interact with the environment. {To help build language-driven embodied agents, various benchmarks based on predefined tasks, datasets and evaluation metrics have been proposed \cite{anderson2018vision, shridhar2020alfred}{, allowing fair comparisons between different approaches}. In these benchmarks, the agent is often given an instruction, following which it is supposed to execute the corresponding sequence of actions.} Even with natural language instructions, such tasks are often overwhelming for the agent \textit{on its own} due to two major challenges: 1) resolving ambiguities in natural language and grounding instructions to actions in a rich environment, and 2) planning for long-horizon action sequences and recovering from possible failures.

\begin{figure}
    \centering
    \includegraphics[width=\columnwidth]{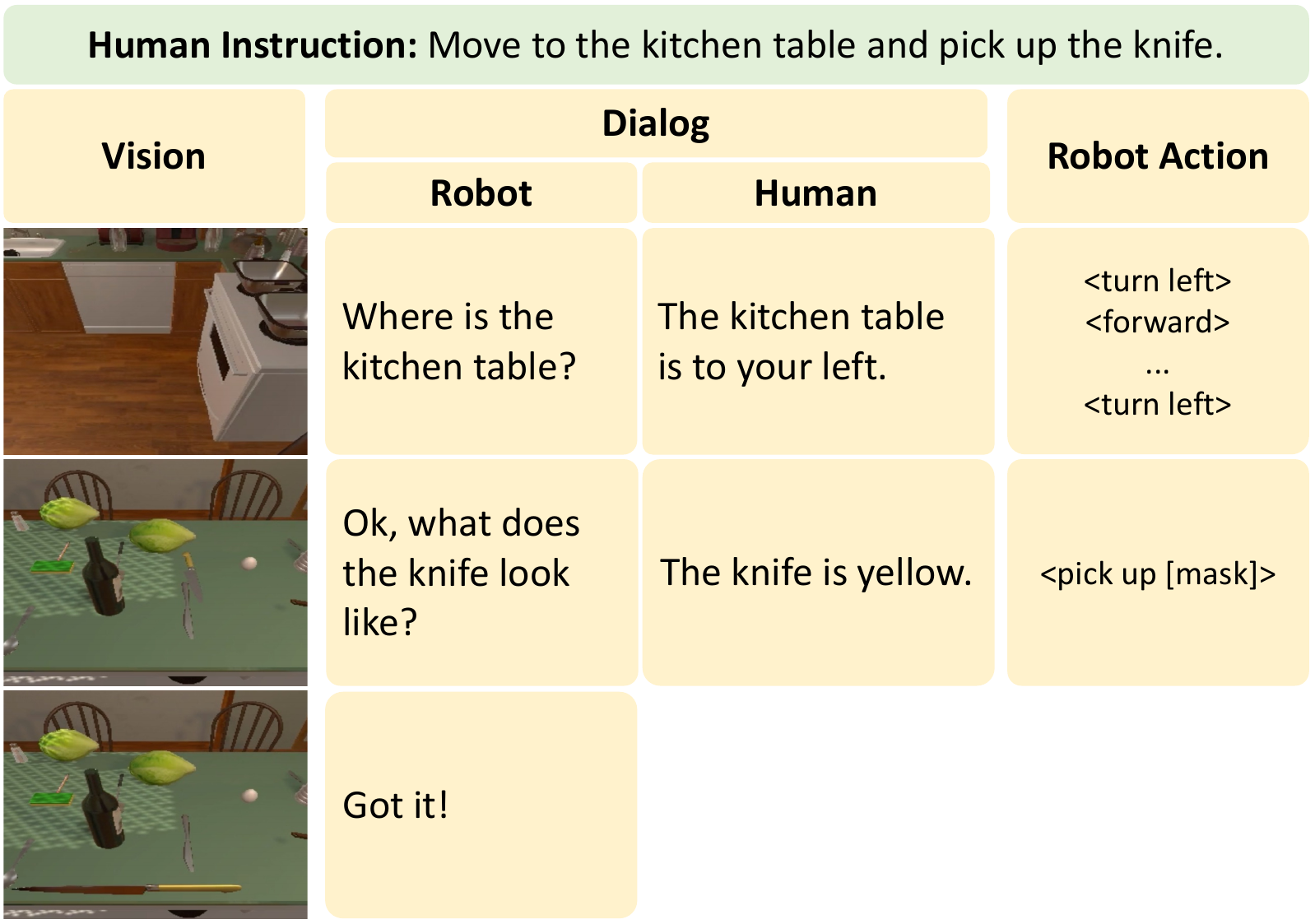}
    %\vspace{-0.1in}
    \caption{\textbf{Example dialogue between a robot and a human user during task completion.} The robot raises questions to obtain additional information (e.g., when the target location is not clear) and to resolve ambiguities (e.g., when facing two knives on the table).}
    \label{fig:figure_1}
    %\vspace{-0.25in}
\end{figure}

Humans, faced with inadequate information for a task, 
seek assistance from others. Similarly, embodied agents should be able to actively ask questions to humans, and utilize the verbal response to overcome challenges in understanding intent and task execution. For example, to deal with ambiguity in human instruction, clarifications are often necessary. As shown in \Cref{fig:figure_1}, the instruction "pick up the knife," is ambiguous when there are two knives in front of the robot -- knowing the color of the intended knife helps the agent ground the instruction to its environment.

We present \textbf{\dialfred}, an embodied instruction following benchmark allowing an agent to 1) actively ask questions to the human user, and 2) use the information in the response to better complete the task. \dialfred\ is built by augmenting \alfred\ \cite{shridhar2020alfred}, an existing benchmark that pairs demonstrations of common household tasks with instructions. \alfred\ language instructions are given as high level goals, e.g., \textquote{Move a knife to the sink}, and a sequence of step-by-step instructions (sub-goals), e.g., \textquote{Move forward to the center table}, \textquote{Pick up the knife}, \textquote{Walk to the sink}, \textquote{Put the knife in the sink}. \alfred\ only contains 7 types of high level goals and 8 types of sub-goals. Existing work \cite{blukis2021persistent, min2022film} has exploited patterns in \alfred\ task structures, and shown that models can acheive state-of-the-art performance by classifying the task type from high-level task instructions alone, even without using step-by-step instructions. To mitigate this issue and ensure the necessity of instruction following, we build \dialfred\ by augmenting \alfred\ for an increased number of task types. In addition, \dialfred\ facilitates agent-human dialogue by providing human-annotated task-relevant questions and answers.

\begin{figure*}
    \centering
    \includegraphics[width=0.95 \textwidth]{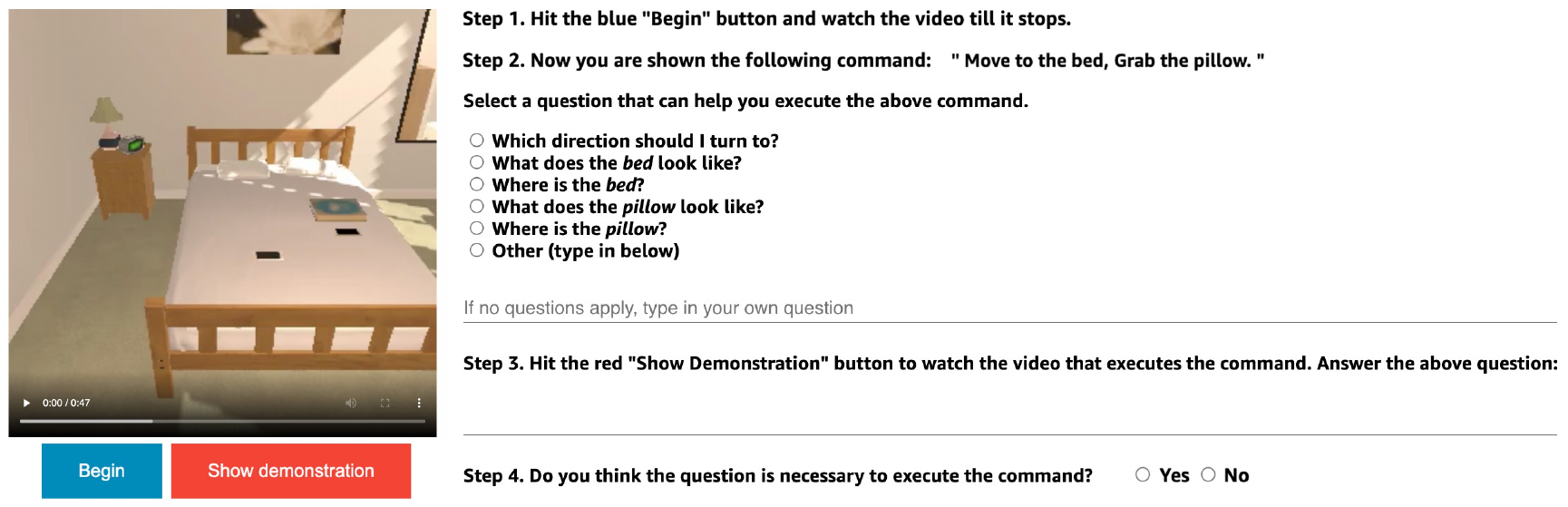}
    \caption{\textbf{The annotation interface for hybrid data collection.} The worker first clicks the ``begin'' button to watch a video clip showing the initial states of the environment. Given the instruction, the worker selects a question to help perform the task. Next, the worker clicks the ``show demonstration'' button to watch the expert demonstration on how to complete the task. The worker then answers their own question based on what they have learned from the videos. Finally, workers choose whether they think the questions and answers are necessary to help the agent carry out the command. }
    \label{fig:hdc_ui}
    \vspace{-15pt}
\end{figure*}

\noindent \textbf{Contributions.} To enable the development and evaluation of dialogue-enabled agents in complex manipulation and navigation tasks, \dialfred\ consists of a) \textbf{25} types of sub-goal level tasks, compared to 8 sub-goals originally available in \alfred; b) \textbf{53K} human-annotated task-relevant questions and answers; and c) models for a questioner-performer framework showing that adding dialogue helps to significantly improve the instruction following performance. We make \dialfred\ publicly available and encourage researchers from related robotics disciplines to propose and evaluate their solutions to dialog-enabled embodied agents.

\section{Related Work}

\noindent \textbf{Embodied Question Answering and Instruction Following.}
Various virtual environments have been developed to accommodate robot agents completing common household tasks~\cite{kolve2017ai2, VRKitchen, xia2020interactive}. Building on these environments, tasks and benchmarks that require an interactive agent to extract information from the environment to answer specific questions have been proposed \cite{embodiedqa, gordon2018iqa}. The other line of work that uses these environments focuses on creating agents to interpret natural language instructions and perform tasks in the environment \cite{anderson2018vision}. \alfred\ \cite{shridhar2020alfred}, a recently proposed benchmark along this direction, requires the agent to complete complex household tasks by following natural language instructions. Dialogue-enabled agents in navigation or manipulation tasks have recently been proposed \cite{thomason2020vision, padmakumar2021teach} -- these focus on action prediction from dialogue history, and do not emphasize the agent's ability to ask clarification questions for better instruction grounding \cite{jokinen2009spoken}. In this paper, we take a further step in this direction by presenting a benchmark for the agent to actively ask questions and learn from the answers to better finish tasks that require navigation and object interactions. We refer readers to this survey \cite{gu2022vision} for dataset comparison. 

\noindent \textbf{Task-Oriented Dialogue.} In task-oriented dialogue, agents rely on skills beyond language modeling (e.g., processing multi-modal sensory data, querying knowledge bases, reasoning based on observations and knowledge \cite{rus2010first,gao2018neural,chai2018language,thomason2020jointly}). Towards building robust dialogue systems, both data-driven \cite{mrkvsic2016neural,hosseini2020simple} and reinforcement learning approaches \cite{su2016line,peng2017composite,hu2020interactive} have been studied. Studies have shown that the ability to ask for help from humans is crucial for agent failure recovery \cite{tellex2014asking}. For visual language navigation, multiple works study when and how to ask for help \cite{nguyen2019vision, nguyen2019help, roman2020rmm, chi2020just}. Household tasks however, pose greater challenges to agents compared to navigation-only tasks, due to longer action sequences, compositional task structures and irreversible object state changes. Our benchmark focuses on these complex household tasks requiring both navigation and manipulation. 

\begin{figure*}[h]
    \centering
    \includegraphics[width=\textwidth]{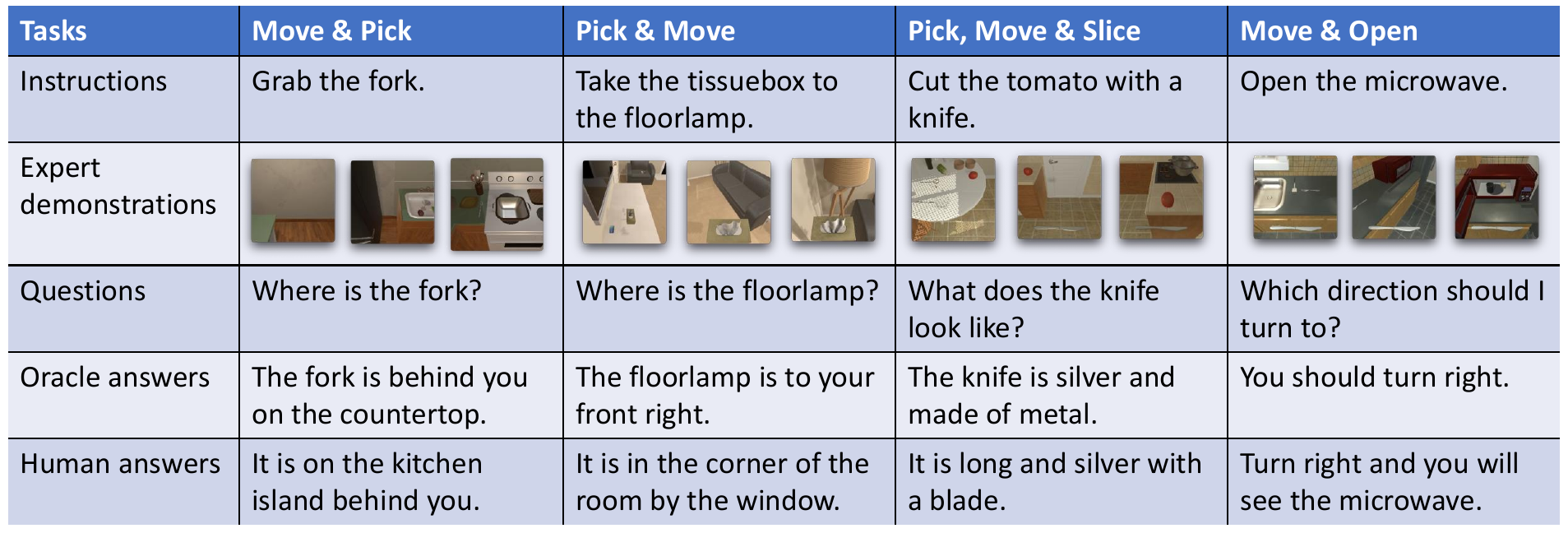}
    %\vspace{-20pt}
    \caption{\textbf{Examples in our QA dataset.} We show instructions and questions asked by humans, and answers provided by both the oracle and humans. Compared to step-by-step instructions, our augmented instructions are concise and general, thus requiring the agent to understand its current state to generate the correct action sequences.}
    \label{fig:task_exp}
    \vspace{-15pt}
\end{figure*}

\section{Task and Dataset}
\dialfred\ requires an agent to follow natural language instructions and perform navigation and object manipulation to finish a household task in a virtual environment. We further enable the agent to ask questions, and use the extra information in the responses to better complete the task.

Each task instance in \dialfred\ is a tuple of the initial environment state, the target environment state and an instruction. The agent's goal is to perform a sequence of actions to change the environment states to the target. Given a natural language instruction, the agent can choose to ask questions to the human, or to execute physical actions in its environment based on the information in the original instructions together with the questions and answers. Physical actions include all 5 navigation actions (e.g., \action{Turn Left}) and all 7 manipulation actions (e.g., \action{Pickup}). These actions can change environment states, and some of the changes are irreversible (e.g. \action{Slice}). Instead of always emitting a physical action, a dialog-enabled agent may emit a question. To standardize this benchmark, we provide an oracle that can answer a set of predefined types of questions (a crude `simulated' human user). The oracle has access to the ground-truth states of the virtual environment, allowing it to provide accurate information regarding objects and tasks. 
%\vspace{-3pt}

\subsection{Hybrid data collection}
\label{sec:hdc}
We collect human annotations on Amazon Mechanical Turk for questions and answers. Each instance in the dataset is a tuple of question type $q \in Q$ asking about a specific property of an object $o \in O$ ($o$ could be empty for questions not related to objects) and a human answer $a \in A$ for the question at the beginning of the task. \Cref{fig:hdc_ui} shows the data collection interface. The hybrid data collection (HDC) process is as follows: 

\begin{enumerate}[leftmargin=*]
\setlength\itemsep{0.5em}
    %\vspace{-5pt}
    \item Each annotator watches a 10 second video clip, which displays the state of the environment right before the task. Annotators also see the original language instructions for the task.
    \item The annotator then selects one pertinent question (from several predefined questions) that they think may help the agent complete the task. The annotator may also type in a question of their own choosing if none of the provided questions are a good fit.
    \item The annotator watches a second video clip -- this time of an expert agent performing the task.
    \item The annotator answers their own question and provides feedback (yes/no) on whether asking the question was necessary in the given scenario. 
\end{enumerate}
\vspace{2pt}

To generate the predefined question choices for the annotator to choose from, we consider three types of questions $Q$, related to the location and appearance of the query object $o$ that needs to be interacted with to finish the task, and the relative direction between the agent's current position and the target position to guide the navigation:
\begin{enumerate}[leftmargin=*]
\setlength\itemsep{0.5em}
    %\vspace{-5pt}
    \item Location: where is $o$?
    \item Appearance: what does $o$ look like?
    \item Direction: which direction should I turn to?
\end{enumerate}
\vspace{2pt}

Given a natural language instruction (e.g. \textquote{Put the egg in the microwave}), we parse it and extract all the nouns (e.g. \textquote{egg} and \textquote{microwave}). We insert each noun into one of the question templates to generate questions (e.g. \textquote{Where is the egg?} and \textquote{What does the microwave look like?}). 

\iffalse
    \begin{table*}[]
    \centering
    \begin{tabular}{|c|c|c|}
    \hline
    sub-goal & low level actions                              &  augmented instructions           \\ \hline
    1              & go to loc 1 + verb noun                       & verb noun                      \\ \hline
    2              & open object + put object + close object       & put object                     \\ \hline
    3              & open object + pick up object + close object   & pick up object                 \\ \hline
    4              & pick up object1 + go to loc1                  & take object1 to loc1           \\ \hline
    5              & turn on object + turn off object              & turn on the object for a while \\ \hline
    6              & pick up object + go to loc2 + put down object & move object to loc2  \\ \hline
    7              & pick up knife + go to object + slice object   & cut the object with a knife    \\ \hline
    \end{tabular}
    \caption{\textbf{Rules for merging low level actions into sub-goals and generating augmented instructions.} Compared to step-by-step instructions, our augmented instructions are concise and general, thus requiring the agent to understand its current state to generate the correct action sequences.}
    \label{tab:data_aug}
    \end{table*}
\fi

We collected human questions and answers for 29,376 sub-goals (e.g., \textquote{Take the knife to the counter}) {on Amazon Mechanical Turk via crowd-sourcing. For each sub-goal, we ask two different annotators to provide questions and answers. And we see a modest level of agreement in question selection between different annotators (Fleiss' $\kappa=0.13$).} To ensure the quality of the dataset, we first remove invalid annotations when the annotation time is less than 15 seconds. A worker is compensated \$0.25 for each {Human Intelligence Task (HIT)}; the dataset collection cost is $\sim$ \$10K. The dataset is gathered over 112 rooms and 80 types of objects. Each human answer contains 6.73 words on average. A lexical complexity analysis on the human answers \cite{lu2012relationship} shows that the number of different words (NDW) in the answers are 7915. The lexical sophistication (the proportion of words not in the 2000 most frequent words in the American National Corpus) is 49\%. 
Example human questions and answers are shown in \Cref{fig:task_exp}.

\begin{figure*}
    \centering
    \includegraphics[width=0.95 \textwidth]{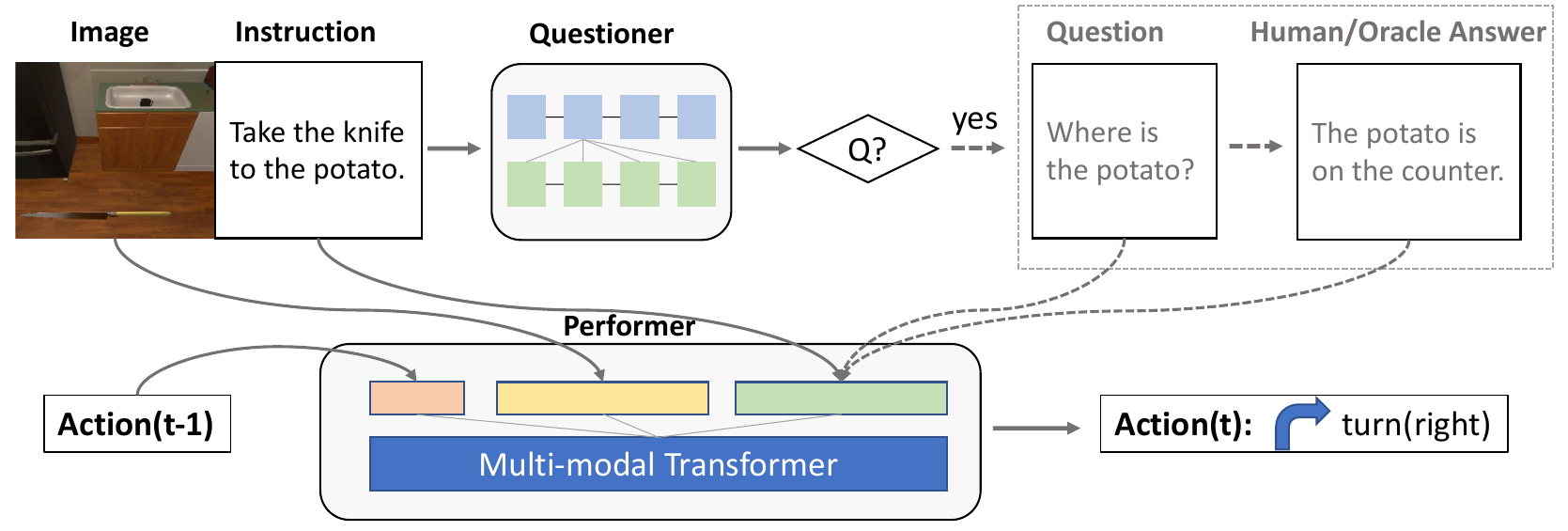}
    \caption{\textbf{The questioner-performer architecture.} The questioner generates questions based on the first person image of the agent and the task instruction. The oracle answers the question based on the scene metadata. The performer takes the image, the instruction, and question and answers as input to predict actions.}
    \label{fig:architecture}
    \vspace{-15pt}
\end{figure*}

\subsection{Generating answers}
In addition to the human answers, we build an oracle that provides templated answers which can be easier for the agent to understand. In \Cref{fig:task_exp}, we show answers generated by the oracle for some task examples. To create the oracle, we take advantage of the ground-truth states of objects and the agent in the simulated environment: (i) to answer object location questions, we compute the direction of the object relative to the agent, and the receptacle that contains the object; (ii) to answer object appearance questions, we focus on its color and material. Object material is extracted from scene metadata in the underlying simulation environment (AI2-Thor). For object color, we extract object pixel RGB values from images, and map them to color names; (iii) for direction questions, we check the agent's location at the end of the task in the ground-truth action sequences, and compare it to the agent's initial location. Based on these metadata, we use language templates to generate answers. Some example templates include: 
\begin{itemize}[leftmargin=*]
\setlength\itemsep{0.5em}
    %\vspace{-5pt}
    \item Location: The $o$ is to your [\textit{direction}] in/on the [\textit{container}].
    \item Appearance: The $o$ is [\textit{color}] and made of [\textit{material}].
    \item Direction: You should turn [\textit{direction}] / You don't need to move.
\end{itemize}

\subsection{Data augmentation on \alfred}

Each task in \alfred\ has a goal, split into multiple sub-goals. Each sub-goal requires the agent to manipulate some objects or move to a target location. Two types of language instructions are given. The high-level task instruction (e.g., \textquote{Move a knife to the sink}) describes the overall goal. The step-by-step instructions (e.g., \textquote{Pick up the knife}) guide the agent to complete each sub-goal. \alfred\ exhibits clear patterns in both high-level task structures and sub-goal action sequences: tasks of the same type require very similar sub-goal sequences to complete them, and sub-goals of the same type (especially manipulation sub-goals) have almost fixed action sequences. The limited variety in tasks and sub-goals precludes instructions understanding -- allowing models that directly classify the task type from high-level task instructions alone to perform well, without even using the  step-by-step instructions \cite{blukis2021persistent, min2022film}. 

To get rid of the strong patterns in both high-level task structures and sub-goal actions in \alfred, \dialfred\ uses data augmentation to increase the number of task types, and focuses on instruction following at the sub-goal level to encourage learning from instructions for each task. We also introduce augmentations on the original \alfred\ sub-goal instructions to add ambiguities in language so that the sequence of actions cannot be fully determined by only focusing on the instruction -- reasoning based on knowledge of the environment state is needed. We show examples for \dialfred\ tasks and instructions in \Cref{fig:task_exp}. 

\noindent \textbf{Sub-goal augmentations.} In \alfred\ there are 8 only sub-goal types (\textquote{go to, pick up, put, cool, heat, clean, slice, toggle}). The action sequences required to finish these are almost fixed. For example, \textquote{cool object} always corresponds to: \textquote{open fridge, put object into fridge, close fridge, open fridge, take out object.} To increase task variations, we augment the sub-goals in two ways. First, we split an original sub-goal into multiple low level actions, each action corresponding to a new sub-goal. For example, the sub-goal \textquote{clean object} is split into three sub-goals: \textquote{put the object in the sink}, \textquote{turn on the faucet} and \textquote{turn off the faucet.} Second, we merge multiple sub-goals into a new sub-goal. For example, sub-goals \textquote{go to the fridge} and \textquote{open the fridge} are merged into a new sub-goal which requires the agent to first go to the fridge and open it. Using these operations, we arrive at 25 sub-goal types in our augmented dataset. In our experiments, we evaluate agent performance on these sub-goals; henceforth referred to as \textbf{tasks}. To standardize the benchmark, we divide the sub-goal task instances into training and validation folds. We further divide the validation fold into \textit{seen} and \textit{unseen} splits depending on whether the environment presents in the training fold. This results in 34,253 tasks in the training fold, 1,296 tasks in the validation seen fold and 1,363 tasks in the validation unseen fold. The corresponding low level actions for all \dialfred\ tasks is displayed in \Cref{fig:25task_info}.

\noindent \textbf{Instruction augmentation.} To generate instructions for \dialfred\ tasks, we create instruction templates for each low level action. For tasks created from split sub-goals, we directly use the template as instruction. For tasks created from combined sub-goals, we concatenate the instructions of low level actions within the sub-goal. In addition to the step-by-step instructions describing low level actions, we generate new instructions that only describe one major action in the task. For example, for the task \textquote{go to the microwave and open the microwave} (\Cref{fig:task_exp}), a human may only describe the main action "open the microwave" in the instruction. The agent cannot determine the action sequence solely based on the instruction. It needs to have an understanding of its position in the room to decide which action to take next. We belive these instructions match human commands in real-world scenrios and are more challenging than the original step-by-step instructions. The example instructions for all \dialfred\ tasks is displayed in \Cref{fig:25task_info}.

\section{Method}

Our baseline for the \textbf{\dialfred} benchmark has two key components, a questioner and a performer (\Cref{fig:architecture}). The questioner asks questions based on the task instruction and agent observations. The performer predicts a sequence of actions to execute in the environment based on the original instruction and the questions and answers. A good questioner knows both when to ask a question and what question(s) to ask, so that the task can be better completed by the performer. We first train the questioner using human-annotated data, i.e., give the model a good starting point by mimicking human judgement. To improve the coordination between the questioner and the performer, we fine-tune the questioner with reinforcement learning. The questions and answers (QAs) together form a dialogue between two entities: the agent (represented by the performer and questioner model) and the human (represented by the instruction and the answers).

\subsection{Architecture}
\label{sec:architecture}

\begin{figure}[t]
    \centering
    \includegraphics[width=\columnwidth]{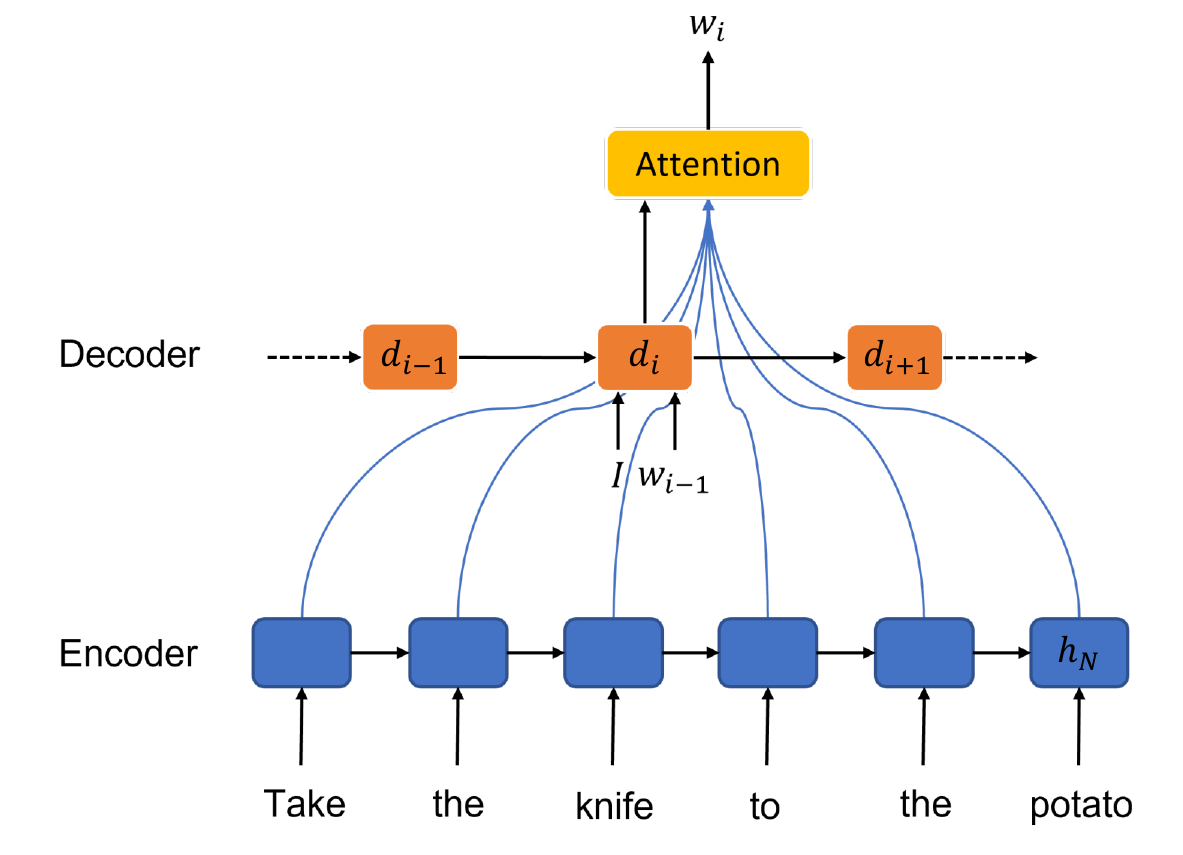}
    \caption{{\textbf{The Questioner model.} Given the instruction and image feature $I$, our model generates question tokens $w_{1:i}$.}}
    \label{fig:questioner}
    %\vspace{-10pt}
\end{figure}

\noindent \textbf{Questioner.}
{Our questioner model (\Cref{fig:questioner}) is based on a sequence-to-sequence architecture with attention \cite{luong2015effective}. It has two modules: an encoder and a decoder, both implemented using LSTM \cite{hochreiter1997long}. The instructions are first tokenized and go through an embedding layer to produce token embeddings $e_{1:N}$. At each time step, the LSTM encoder takes the embedding of the current token $e_{n}$ and the previous encoder hidden state $h_{n-1}$ to produce a new hidden state $h_{n}$. As a result, the encoder generates a sequence of hidden states $h_{1:N}$. The final encoder hidden state $h_{N}$ is used to initialize the decoder's hidden state $d_{0}$. The decoder uses the image ResNet feature $I$, the previous question token $w_{i-1}$ and the previous decoder hidden state $d_{i-1}$ to produce the new docoder hidden state $d_{i}$. The decoder hidden state $d_{i}$ is used to attend over encoder hidden states $h_{1:N}$ and predict the next question token $w_{i}$ via a dot-product attention mechanism. Each question is represented by two tokens, including one for the question type (i.e. \textit{loc} for location, \textit{app} for appearance and \textit{dir} for direction), and the other for the target object. The questioner can also choose not to ask a question by generating a \textit{none} token. We pre-train the questioner using questions selected by Turkers (\Cref{sec:hdc}) on the training split. }

%\vspace{-5pt}
\noindent \textbf{Performer.} Our performer is based on the Episodic Transformer \cite{pashevich2021episodic}, an attention-based multi-layer transformer model that encodes the full history of the instruction and QAs, visual observations and action history to predict future actions. To enable the model to handle all possible QAs from the questioner and oracle, we pre-train it on the training split by providing it with instructions and all possible questions, including a combination of question types and answers. Model parameters are optimized by minimizing the cross-entropy between predicted and expert actions.
%\vspace{-3pt}

% \noindent \textbf{RL fine-tuning.}
\subsection{Questioner fine-tuning using RL }
The goal of the questioner is to ask necessary questions to help the performer finish the task. Thus we fine-tune the questioner using reinforcement learning to learn {\em when} to ask a question and {\em what} question(s) to ask on the validation seen split. The learning system for question asking is modeled as a Markov Decision Process, specified by a tuple $<S, A, T, R>$, where $S$ is the state space {(including the language instructions and visual observations)}, $A=Q \times O$ is the action space of all possible questions (in our cases a combination of question types $Q$ and target object $O$),  $T(s'|s, a)$ is the transition function encoding how the performer can advance the task given the question and its corresponding answers, and $R(s,a)$ is the reward function encoding the reward for each $(s,a)$ pair. {Our questioner model can be viewed as a policy network $\pi_{\theta}(s,a)=p(a|s;\theta)$ mapping each state vector $s$ to a stochastic questioning policy, which can be learned based on the reward function to strike a balance between the number of questions and performance gain.}

{We adopt the following reward function structure to address this trade-off: reward for task completion $r_{suc}=1.0$, penalty for each step $r_{step}=-0.01$, penalty per question asked $r_{q}=-0.05$, penalty per invalid question $r_{invalid}=-0.1$. Some questions generated by the questioner cannot be answered by the oracle, e.g. the appearance of task-irrelevant objects. Thus we add a penalty for invalid questions. The reward forms the actor critic loss function \cite{sutton2018reinforcement}, the gradient of which is used to update the model so that it can learn to ask necessary questions at correct time. The performer model is not updated during questioner fine-tuning.}

\begin{table*}[h]
\centering
\begin{tabular}{|c|l|c|c|c|c|c|}
\hline
\# & Expt setting  & Seen SR & Unseen SR & Seen PWSR & Unseen PWSR  & NQ  \\ \hline
1   & Instruction only   & 25.4   & 18.3  & 18.4  &  11.4  &  0  \\ \hline
2   & All QAs & 43.4   & 32.0   & 31.2 & 19.9 & 3.24  \\ \hline
3   & Random QA & 39.9  & 27.9 & 28.4  &  17.3 &  0.81 \\ \hline
4   & Random MC  & 46.6  & 29.5  & \textbf{35.5} & 18.7 & 0.52  \\ \hline
5   & RL begin & 47.3  & 32.7  & 33.5 &  20.1 &  0.37 \\ \hline
6   & RL anytime &  \textbf{47.8}  &  \textbf{33.6}  & 34.2 & \textbf{20.4}  &  0.71 \\ \hline

\end{tabular}
\caption{\textbf{Performance of the baselines.} Seen SR and unseen SR represent the success rate on valid seen and valid unseen splits. PWSR is the path weighted success rate. NQ is the number of questions asked by the questioner. The best results are highlighted in boldface.}
\label{tab:baselines}
%\vspace{-5pt}
\end{table*}

\begin{table*}[h]
    \centering
    \begin{tabular}{|l|c|c|c|c|c|c|}
    \hline
    Expt setting & Unseen SR & Unseen PWSR & NQ & Loc Perc. & App Perc. & Dir Perc. \\ \hline
    Human & -  & - & 0.66 & 0.72 & 0.22 & 0.06  \\ \hline
    RL anytime & 33.6 & 20.4 & 0.71 & 0.65 & 0.14 & 0.21 \\ \hline
    RL Loc perturb & 32.2 & 19.8 & 0.66 & 0.47 & 0.15 & 0.38 \\ \hline
    RL App perturb & 32.4 & 19.9 & 0.40 & 0.92 & 0.04 & 0.04 \\ \hline
    RL Dir perturb & 33.0 & 20.4 & 0.61 & 0.51 & 0.45 & 0.04 \\ \hline
    Random  & 28.4  &  17.3 & 0.81 & 0.36 & 0.31 & 0.33 \\ \hline
    Random Loc perturb & 26.0 & 16.0 & 0.81 & 0.36 & 0.31 & 0.33 \\ \hline
    Random App perturb & 26.1 & 16.1 & 0.81 & 0.36 & 0.31 & 0.33 \\ \hline
    Random Dir perturb & 26.2 & 15.8 & 0.81 & 0.36 & 0.31 & 0.33 \\ \hline
    \end{tabular}
    \caption{\textbf{Ablation study by perturbing the oracle.} We start from two settings: a questioner that has been fine-tuned for asking questions at any time and a questioner that asks a random question at the beginning. We perturb the oracle by not providing answers for one question type 50\% of the time. Loc Perc, App Perc and Dir Perc represent the percentage questions about object locations, object appearance and directions respectively. }
    \label{tab:ablation_qtype}
    %\vspace{-15pt}
\end{table*}

\begin{table*}[h]
    \centering
    \begin{tabular}{|l|c|c|c|c|c|}
    \hline
    Experiment setting  & Seen SR & Unseen SR & Seen PWSR & Unseen PWSR  & NQ  \\ \hline
    % RL anytime (Fixed 1) &  47.1  & 33.2  & 33.2 & 19.9  &  1.27 \\ \hline
    % RL anytime (Fixed 1) &  44.2  & 30.1  &  31.7  & 19.3  &  0.39 \\ \hline
    RL anytime (Fixed 1) & \textbf{51.9} & \textbf{34.7} & \textbf{36.3} & \textbf{21.2} & 21.39 \\ \hline
    RL anytime (Fixed 5) &  47.8  &  {33.6}  & 34.2 & 20.4  &  0.71 \\ \hline
    RL anytime (Fixed 10) &  46.3  &  {32.4}  &  32.7  & {19.9}  &  0.36 \\ \hline
    RL anytime (MC) &  47.1  & 33.0  & 33.3  & 20.4  & 0.31 \\ \hline
    
    \end{tabular}
    \caption{\textbf{Effect of question timing.} We manipulate the number of steps the performer rolls out before the questioner can ask the next question. For \textit{(Fixed 1)}, we modify the rewards ($r_{invalid}=-0.01, r_{q}=-0.002$) to promote question asking. For \textit{(MC)}, the questioner asks questions based on the performer model confusion.}
    \label{tab:ablation_timing}
    \vspace{-15pt}
\end{table*}

\subsection{Heuristic-based questioner}
\label{sec:heuristic}
Inspired by \cite{chi2020just}, we implemented a questioner based on model confusion. The idea is that if the performer is not confident, the output action distribution would have high entropy; model confusion could be a good heuristic to know when to ask a question. We sort action probabilities in decreasing order; an agent is confused if the minimum difference between the top two actions is less than a threshold $\epsilon$ throughout the  action sequences:
\begin{equation}
    \min_{t} (p_{sorted}^{t}[0] - p_{sorted}^{t}[1]) < \epsilon
    \label{eq:mc}
\end{equation}
where the threshold $\epsilon$ is used to control the degree of confusion for asking questions. 
% The MC questioner randomly asks a question when the performer is confused. 
In practice, we set the confusion threshold $\epsilon=0.5$ in the experiment. 

\section{Experiments}
We evaluate the baseline models on our dataset. We terminate the episode when it exceeds 1000 steps or has more than 10 failed actions.

\subsection{Evaluation metrics}
We evaluate model performance using success rate and path weighted success rate. {In \dialfred, task success $s$ is a binary indicator of whether the task goal conditions have been achieved. For example, the task ``clean the fork'' requires the states of the fork to be changed from ``dirty'' to ``clean''. Task success $s$ is defined as 1 if the goal conditions have been fully achieved, and 0 otherwise. Success rate is calculated via averaging $s$ across all tasks.
Path weighted success rate further takes into consideration the number of actions the robot takes. The path weighted success score for a task is calculated by:
\begin{equation}
    p_{s} = s * L^{*} / \max(L^{*}, \hat{L})
\end{equation}
where $s$ is the original task success indicator (0 or 1), $\hat{L}$ is the number of steps taken by the agent to complete the task, and $L^{*}$ is the number of steps in the expert demonstration. Intuitively, the more steps the robot took, the lower the score. Again, the path weighted success rate is calculated via averaging $p_{s}$ across all tasks.} To understand the trade-off between the number of questions asked and task performance, we measure the number of questions throughout the task. 

\subsection{Baselines}
We implemented 6 baselines and evaluated their performance on our augmented dataset (\Cref{tab:baselines}). In all baselines we use the Episodic Transformer model as the performer. In baselines 2--6, the performer is trained using imitation learning on expert action sequences with language instructions and QAs. In baselines 5--6, the questioner is pre-trained on human dialogues in the training split, and fine-tuned using reinforcement learning on the valid seen split. Baseline details are itemized below:
%\vspace{-5pt}
\begin{enumerate}[leftmargin=*]
\setlength\itemsep{0.2em}
    \item In this baseline, no questions are allowed, the performer is trained to predict the action sequences based on the language instruction and visual observations.
    \item In addition to the instruction, the performer gets all valid QAs at the beginning of the task, including a combination of all three question types (i.e. location, appearance and direction) and objects mentioned in the instruction. 
    \item Questions are sampled randomly based on type: 25\% for each of the 3 types and 25\% for no question. Given a selected question type, questions and answers for all relevant objects are given to the performer in addition to the instructions.
    \item When the action sequences generated by the performer satisfy the model confusion criterion (\Cref{eq:mc}), the performer is randomly provided with a valid QA as input.
    \item The questioner is fine-tuned using reinforcement learning to learn whether to ask a question and what question to ask at the beginning of a task. The performer uses the QA to generate the action sequence to finish the task. 
    \item Similar to baseline 5, the questioner is fine-tuned using reinforcement learning, but now it can ask questions in the middle of the task. Given the instruction and previous QAs, we roll out the performer for 5 steps, following which the questioner is allowed to generate new questions and thus get new answers.
\end{enumerate}

%\vspace{-10pt}
\subsection{Results}
We display (\Cref{tab:baselines}) the task performance (success rate) on validation seen and unseen splits for all baseline models. Comparing the results of baselines 1--3, we  see that adding QAs to the instructions improves task performance on both splits. Comparing the results of baselines 2--5, we see that the fine-tuned questioner achieves the best performance, with a smaller number of questions. Comparing the results of two reinforcement learning baselines (5,6), we see that enabling the agent to ask questions in the middle of the task improves performance at the cost of more questions and answers. In addition, the \textit{random MC} baseline achieves reasonable performance on the valid seen split, but not on the valid unseen split. Since the performer is pre-trained on the training split, which has the same scene as the valid seen split, and is given random combinations of different types of QAs as input, it is not surprising that the \textit{random MC} baseline does not generalize well to unseen environments.

\subsection{Ablation Study}

\noindent \textbf{Perturbed oracle.} We perform an ablation study (\Cref{tab:ablation_qtype}) by perturbing the oracle. For each question type, we limit the oracle to provide answers for 50\% of the asked questions. The original fine-tuned questioner (\textit{RL anytime}) has a similar number of questions (and distribution) as the human data it is pre-trained on -- most questions are about object locations. With the perturbation, the model asks slightly fewer questions. We observe that the fine-tuned questioner model adapts to the deficient oracle. For example, after perturbation on object location answers, the questioner asks 28\% fewer location questions and 81\% more direction questions instead. Looking at model performance after perturbation, we find that perturbing the location answers reduces the performance the most. This result matches the large proportion of location questions asked by both humans and the learning-tuned model, indicating that location answers are probably most useful for task completion. We perform the same perturbations on the oracle for the \textit{random} questioner. Comparing the performance of models before and after perturbation, we see that the drop in SR caused by the perturbations for the learning-tuned questioners are smaller than the drops for the \textit{random} questioners. The difference can be explained by the learning-tuned questioner's ability to adjust the proportion of questions to adapt to the deficient oracle.

\noindent \textbf{Question timing.} To understand how question timing affects performance, we change the number of action steps the performer executes before allowing the questioner to ask a question. We add a setting, \textquote{RL anytime (MC)}, which allows the questioner to ask questions based on model confusion (\Cref{eq:mc}) during fine-tuning. The results (\Cref{tab:ablation_timing}) show that reducing the number of steps between questions leads to slightly better performance, but it also requires the oracle to answer significantly more questions. Comparing the results of model confusion with fixed timing, we find that it achieves reasonable performance at relatively low cost.

\section{Conclusion}

We presented \textbf{\dialfred}, a dialogue-enabled embodied instruction following benchmark that allows an agent to actively ask questions while interacting with the environment to finish a household task. \dialfred\ is generated by augmenting \alfred\ to increase task and language variations. It includes an oracle to answer questions and a human-annotated dataset with 53K task-relevant questions and answers -- a potential resource to model how humans ask and answer task-oriented questions. To tackle \dialfred, we propose a questioner-performer baseline (and variants) wherein the questioner is pre-trained with the human-annotated data and fine-tuned with reinforcement learning.
Experimental results show that asking the right questions leads to significantly improved task performance. Extending existing embodied instruction following benchmarks with dialogue is a promising avenue of research towards truly interactive embodied agents.Along these lines, we posit that the general framework of oracle-guided reinforcement training and the hybrid data annotation method we employ may be useful to ``dialogue-enable'' other embodied instruction following tasks.

\begin{table*}[h]
\centering
\begin{tabular}{|l|l|l|l|}
\hline
\textbf{\#} & \textbf{Tasks}        & \textbf{Example Low Level Actions}               & \textbf{Example Instructions}        \\ \hline
1           & Clean                 & \begin{tabular}[c]{@{}l@{}}put sink, turnon faucet, turnoff faucet, pickup pot\end{tabular}                                                                                                          & Wash the pot.                        \\ \hline
2           & Close                 & close drawer                                                                                                                                                                                           & Close the drawer.                    \\ \hline
3           & Cool                  & \begin{tabular}[c]{@{}l@{}}open fridge, put fridge, close fridge, open fridge, take tomato, close fridge    \end{tabular}                                                                                                                                                                & Chill the tomato.                    \\ \hline
4          & Heat                  & \begin{tabular}[c]{@{}l@{}}open microwave, put microwave, close microwave, turnon microwave, \\ turnoff microwave, open microwave, pickup apple, close microwave\end{tabular}                    & Cook the apple.                      \\ \hline
5          & Move                  & goto cart                                                                                                                                                                                              & Go to the cart.                      \\ \hline
6          & Open                  & open safe                                                                                                                                                                                              & Open the safe.                       \\ \hline
7          & Pick                  & pickup egg                                                                                                                                                                                             & Take the egg.                        \\ \hline
8          & Put                   & put plate                                                                                                                                                                                              & Put on the plate.                    \\ \hline
9          & Slice                 & slice letttuce                                                                                                                                                                                         & Cut the lettuce.                     \\ \hline
10          & Turn on               & turnon desklamp                                                                                                                                                                                        & Power on the desklamp.               \\ \hline
11          & Turn off              & turnoff microwave                                                                                                                                                                                      & Switch off the microwave.            \\ \hline
12           & Move \& Clean         & \begin{tabular}[c]{@{}l@{}}goto sinkbasin, put sink, turnon faucet, turnoff faucet, pickup ladle\end{tabular}                                                                                       & Go to the sinkbasin, wash the ladle. \\ \hline
13           & Move \& Close         & goto laptop, close laptop                                                                                                                                                                              & Close the laptop.                    \\ \hline
14          & Move \& Cool          & \begin{tabular}[c]{@{}l@{}}goto fridge, open fridge, put fridge, close fridge, open fridge,\\ take potato, close fridge\end{tabular}                                                                                                           & Move to the fridge, cool the potato. \\ \hline
15           & Move \& Heat          & \begin{tabular}[c]{@{}l@{}}goto microwave, open microwave, put microwave, close microwave, \\ turnon microwave, turnoff microwave, open microwave, pickup potato, \\ close microwave\end{tabular} & Heat up the potato.                  \\ \hline
16           & Move \& Open          & goto microwave, open microwave                                                                                                                                                                         & Open the microwave.                  \\ \hline
17           & Move \& Pick          & goto countertop, pickup fork                                                                                                                                                                           & Grab the fork.                       \\ \hline
18          & Move \& Put           & goto countertop, put countertop                                                                                                                                                                        & Place on the countertop.             \\ \hline
19          & Move \& Slice         & goto bread, slice bread                                                                                                                                                                                & Cut the bread.                       \\ \hline
20          & Move \& Turn on       & goto desklamp, turnon desklamp                                                                                                                                                                         & Switch on the desklamp.              \\ \hline
21          & Open \& Pick \& Close & \begin{tabular}[c]{@{}l@{}}open microwave, pickup cup, close microwave\end{tabular}                                                                                                                 & Pick up the cup.                     \\ \hline
22          & Open \& Put \& Close  & open fridge, put fridge, close fridge                                                                                                                                                                  & Put into the fridge.                 \\ \hline
23          & Pick \& Move          & pickup mug, goto sidetable                                                                                                                                                                             & Take the mug to the sidetable.       \\ \hline
24          & Pick \& Move \& Put   & \begin{tabular}[c]{@{}l@{}}pickup keychain, goto armchair, put keychain\end{tabular}                                                                                                                & Put the keychain on the armchair.    \\ \hline
25          & Pick \& Move \& Slice & pickup knife, goto bread, slice bread                                                                                                                                                                  & Cut the bread with a knife           \\ \hline
\end{tabular}
\caption{{List of all 25 tasks in \dialfred\ and their example low level actions and instructions. Note that here we do not further split ``goto'' into low level actions (e.g. move forward, turn left, etc.).}}
\vspace{-18pt}
\label{fig:25task_info}
\end{table*}

\iffalse
\section*{ACKNOWLEDGMENT}

The preferred spelling of the word ÒacknowledgmentÓ in America is without an ÒeÓ after the ÒgÓ. Avoid the stilted expression, ÒOne of us (R. B. G.) thanks . . .Ó  Instead, try ÒR. B. G. thanksÓ. Put sponsor acknowledgments in the unnumbered footnote on the first page.
\fi

%%%%%%%%%%%%%%%%%%%%%%%%%%%%%%%%%%%%%%%%%%%%%%%%%%%%%%%%%%%%%%%%%%%%%%%%%%%%%%%%
\balance
\bibliographystyle{IEEEtran}
\bibliography{IEEEexample}

\begin{thebibliography}{10}
\providecommand{\url}[1]{#1}
\csname url@rmstyle\endcsname
\providecommand{\newblock}{\relax}
\providecommand{\bibinfo}[2]{#2}
\providecommand\BIBentrySTDinterwordspacing{\spaceskip=0pt\relax}
\providecommand\BIBentryALTinterwordstretchfactor{4}
\providecommand\BIBentryALTinterwordspacing{\spaceskip=\fontdimen2\font plus
\BIBentryALTinterwordstretchfactor\fontdimen3\font minus
  \fontdimen4\font\relax}
\providecommand\BIBforeignlanguage[2]{{%
\expandafter\ifx\csname l@#1\endcsname\relax
\typeout{** WARNING: IEEEtran.bst: No hyphenation pattern has been}%
\typeout{** loaded for the language `#1'. Using the pattern for}%
\typeout{** the default language instead.}%
\else
\language=\csname l@#1\endcsname
\fi
#2}}

\bibitem{anderson2018vision}
P.~Anderson, Q.~Wu, D.~Teney, J.~Bruce, M.~Johnson, N.~S{\"u}nderhauf, I.~Reid,
  S.~Gould, and A.~Van Den~Hengel, ``Vision-and-language navigation:
  Interpreting visually-grounded navigation instructions in real
  environments,'' in \emph{Proceedings of the IEEE Conference on Computer
  Vision and Pattern Recognition}, 2018, pp. 3674--3683.

\bibitem{shridhar2020alfred}
M.~Shridhar, J.~Thomason, D.~Gordon, Y.~Bisk, W.~Han, R.~Mottaghi,
  L.~Zettlemoyer, and D.~Fox, ``Alfred: A benchmark for interpreting grounded
  instructions for everyday tasks,'' in \emph{Proceedings of the IEEE/CVF
  conference on computer vision and pattern recognition}, 2020, pp.
  10\,740--10\,749.

\bibitem{blukis2021persistent}
V.~Blukis, C.~Paxton, D.~Fox, A.~Garg, and Y.~Artzi, ``A persistent spatial
  semantic representation for high-level natural language instruction
  execution,'' in \emph{Conference on Robot Learning}, 2021, pp. 706--717.

\bibitem{min2022film}
S.~Y. Min, D.~S. Chaplot, P.~K. Ravikumar, Y.~Bisk, and R.~Salakhutdinov,
  ``{FILM}: Following instructions in language with modular methods,'' in
  \emph{International Conference on Learning Representations}, 2022.

\bibitem{kolve2017ai2}
E.~Kolve, R.~Mottaghi, W.~Han, E.~VanderBilt, L.~Weihs, A.~Herrasti, D.~Gordon,
  Y.~Zhu, A.~Gupta, and A.~Farhadi, ``Ai2-thor: An interactive 3d environment
  for visual ai,'' \emph{arXiv preprint arXiv:1712.05474}, 2017.

\bibitem{VRKitchen}
X.~Gao, R.~Gong, T.~Shu, X.~Xie, S.~Wang, and S.~Zhu, ``Vrkitchen: an
  interactive 3d virtual environment for task-oriented learning,''
  \emph{arXiv}, vol. abs/1903.05757, 2019.

\bibitem{xia2020interactive}
F.~Xia, W.~B. Shen, C.~Li, P.~Kasimbeg, M.~E. Tchapmi, A.~Toshev,
  R.~Mart{\'\i}n-Mart{\'\i}n, and S.~Savarese, ``Interactive gibson benchmark:
  A benchmark for interactive navigation in cluttered environments,''
  \emph{IEEE Robotics and Automation Letters}, vol.~5, no.~2, pp. 713--720,
  2020.

\bibitem{embodiedqa}
A.~Das, S.~Datta, G.~Gkioxari, S.~Lee, D.~Parikh, and D.~Batra, ``{E}mbodied
  {Q}uestion {A}nswering,'' in \emph{Proceedings of the IEEE Conference on
  Computer Vision and Pattern Recognition}, 2018.

\bibitem{gordon2018iqa}
D.~Gordon, A.~Kembhavi, M.~Rastegari, J.~Redmon, D.~Fox, and A.~Farhadi, ``Iqa:
  Visual question answering in interactive environments,'' in \emph{Proceedings
  of the IEEE conference on computer vision and pattern recognition}, 2018, pp.
  4089--4098.

\bibitem{thomason2020vision}
J.~Thomason, M.~Murray, M.~Cakmak, and L.~Zettlemoyer, ``Vision-and-dialog
  navigation,'' in \emph{Conference on Robot Learning}, 2020, pp. 394--406.

\bibitem{padmakumar2021teach}
A.~Padmakumar, J.~Thomason, A.~Shrivastava, P.~Lange, A.~Narayan-Chen,
  S.~Gella, R.~Piramithu, G.~Tur, and D.~Hakkani-Tur, ``{TEACh: Task-driven
  Embodied Agents that Chat},'' in \emph{Conference on Artificial Intelligence
  (AAAI)}, 2022.

\bibitem{jokinen2009spoken}
K.~Jokinen and M.~McTear, ``Spoken dialogue systems,'' \emph{Synthesis Lectures
  on Human Language Technologies}, vol.~2, no.~1, pp. 1--151, 2009.

\bibitem{gu2022vision}
J.~Gu, E.~Stefani, Q.~Wu, J.~Thomason, and X.~E. Wang, ``Vision-and-language
  navigation: A survey of tasks, methods, and future directions,'' \emph{arXiv
  preprint arXiv:2203.12667}, 2022.

\bibitem{rus2010first}
V.~Rus, B.~Wyse, P.~Piwek, M.~Lintean, S.~Stoyanchev, and C.~Moldovan, ``The
  first question generation shared task evaluation challenge,'' in
  \emph{Proceedings of the 6th International Natural Language Generation
  Conference}, 2010.

\bibitem{gao2018neural}
J.~Gao, M.~Galley, and L.~Li, ``Neural approaches to conversational ai,'' in
  \emph{The 41st International ACM SIGIR Conference on Research \& Development
  in Information Retrieval}, 2018, pp. 1371--1374.

\bibitem{chai2018language}
J.~Y. Chai, Q.~Gao, L.~She, S.~Yang, S.~Saba-Sadiya, and G.~Xu, ``Language to
  action: Towards interactive task learning with physical agents.'' in
  \emph{IJCAI}, 2018, pp. 2--9.

\bibitem{thomason2020jointly}
J.~Thomason, A.~Padmakumar, J.~Sinapov, N.~Walker, Y.~Jiang, H.~Yedidsion,
  J.~Hart, P.~Stone, and R.~Mooney, ``Jointly improving parsing and perception
  for natural language commands through human-robot dialog,'' \emph{Journal of
  Artificial Intelligence Research}, vol.~67, pp. 327--374, 2020.

\bibitem{mrkvsic2016neural}
N.~Mrk{\v{s}}i{\'c}, D.~{\'O}~S{\'e}aghdha, T.-H. Wen, B.~Thomson, and
  S.~Young, ``Neural belief tracker: Data-driven dialogue state tracking,'' in
  \emph{Proceedings of the 55th Annual Meeting of the Association for
  Computational Linguistics (Volume 1: Long Papers)}, Vancouver, Canada, July
  2017, pp. 1777--1788.

\bibitem{hosseini2020simple}
E.~Hosseini-Asl, B.~McCann, C.-S. Wu, S.~Yavuz, and R.~Socher, ``A simple
  language model for task-oriented dialogue,'' \emph{arXiv preprint
  arXiv:2005.00796}, 2020.

\bibitem{su2016line}
P.-H. Su, M.~Ga{\v{s}}i{\'c}, N.~Mrk{\v{s}}i{\'c}, L.~M. Rojas-Barahona,
  S.~Ultes, D.~Vandyke, T.-H. Wen, and S.~Young, ``On-line active reward
  learning for policy optimisation in spoken dialogue systems,'' in
  \emph{Proceedings of the 54th Annual Meeting of the Association for
  Computational Linguistics (Volume 1: Long Papers)}, Berlin, Germany, Aug.
  2016, pp. 2431--2441.

\bibitem{peng2017composite}
B.~Peng, X.~Li, L.~Li, J.~Gao, A.~Celikyilmaz, S.~Lee, and K.-F. Wong,
  ``Composite task-completion dialogue policy learning via hierarchical deep
  reinforcement learning,'' in \emph{Proceedings of the Conference on Empirical
  Methods in Natural Language Processing (EMNLP)}, 2017.

\bibitem{hu2020interactive}
X.~Hu, Z.~Wen, Y.~Wang, X.~Li, and G.~de~Melo, ``Interactive question
  clarification in dialogue via reinforcement learning,'' in \emph{Proceedings
  of the 28th International Conference on Computational Linguistics: Industry
  Track}, 2020, pp. 78--89.

\bibitem{tellex2014asking}
S.~Tellex, R.~Knepper, A.~Li, D.~Rus, and N.~Roy, ``Asking for help using
  inverse semantics,'' in \emph{Proceedings of Robotics: Science and Systems},
  Berkeley, USA, July 2014.

\bibitem{nguyen2019vision}
K.~Nguyen, D.~Dey, C.~Brockett, and B.~Dolan, ``Vision-based navigation with
  language-based assistance via imitation learning with indirect
  intervention,'' in \emph{Proceedings of the IEEE/CVF Conference on Computer
  Vision and Pattern Recognition}, 2019, pp. 12\,527--12\,537.

\bibitem{nguyen2019help}
K.~Nguyen and H.~Daum{\'e}~III, ``Help, anna! visual navigation with natural
  multimodal assistance via retrospective curiosity-encouraging imitation
  learning,'' in \emph{Proceedings of the Conference on Empirical Methods in
  Natural Language Processing (EMNLP)}, November 2019.

\bibitem{roman2020rmm}
H.~R. Roman, Y.~Bisk, J.~Thomason, A.~Celikyilmaz, and J.~Gao, ``Rmm: A
  recursive mental model for dialog navigation,'' in \emph{Findings of the 2020
  Conference on Empirical Methods in Natural Language Processing}, 2020.

\bibitem{chi2020just}
T.-C. Chi, M.~Shen, M.~Eric, S.~Kim, and D.~Hakkani-tur, ``Just ask: An
  interactive learning framework for vision and language navigation,'' in
  \emph{Proceedings of the AAAI Conference on Artificial Intelligence},
  vol.~34, no.~03, 2020, pp. 2459--2466.

\bibitem{lu2012relationship}
X.~Lu, ``The relationship of lexical richness to the quality of esl learners’
  oral narratives,'' \emph{The Modern Language Journal}, vol.~96, no.~2, pp.
  190--208, 2012.

\bibitem{luong2015effective}
M.-T. Luong, H.~Pham, and C.~D. Manning, ``Effective approaches to
  attention-based neural machine translation,'' in \emph{Proceedings of the
  2015 Conference on Empirical Methods in Natural Language Processing}, 2015,
  pp. 1412--1421.

\bibitem{hochreiter1997long}
S.~Hochreiter and J.~Schmidhuber, ``Long short-term memory,'' \emph{Neural
  computation}, vol.~9, no.~8, pp. 1735--1780, 1997.

\bibitem{pashevich2021episodic}
A.~Pashevich, C.~Schmid, and C.~Sun, ``Episodic transformer for
  vision-and-language navigation,'' \emph{2021 IEEE/CVF International
  Conference on Computer Vision (ICCV)}, pp. 15\,922--15\,932, 2021.

\bibitem{sutton2018reinforcement}
R.~S. Sutton and A.~G. Barto, \emph{Reinforcement learning: An
  introduction}.\hskip 1em plus 0.5em minus 0.4em\relax MIT press, 2018.

\end{thebibliography}

\end{document}